\documentclass[runningheads]{llncs}
\usepackage{comment}
\usepackage{graphicx}
\usepackage{cite}
\usepackage{colortbl}
\usepackage{indentfirst}
\usepackage{color}
\usepackage[utf8]{inputenc} 
\usepackage[T1]{fontenc}    
\usepackage{url}            
\usepackage{booktabs}       
\usepackage{amsfonts}       
\usepackage{nicefrac}       
\usepackage{microtype}      
\usepackage{authblk}
\usepackage{hyperref}
\usepackage[cmyk]{xcolor}
\begin{document}
%




\title{AIGCIQA2023:  A Large-scale Image Quality Assessment Database for AI Generated Images: from the Perspectives of Quality, Authenticity \\ and Correspondence}
%
\author{Jiarui Wang\inst{1}\and
Huiyu Duan\inst{1}\and
Jing Liu\inst{2}\and
Shi Chen\inst{3}\\
Xiongkuo Min\inst{1}$^*$, and Guangtao Zhai\inst{1}\thanks{Corresponding Authors.}
}

\authorrunning{F. Author et al.}

\institute{Shanghai Jiao Tong University, Shanghai, China \\
\email{\{wangjiarui,huiyuduan,minxiongkuo,zhaiguangtao\}@sjtu.edu.cn}
\and Tianjin University, Tianjin, China\\
\and Shanghai Second Polytechnic University, Shanghai, China\\}
\maketitle              
\begin{abstract}
    
    Recent years have witnessed a rapid growth of Artificial Intelligence Generated Content (AIGC), among which with the development of text-to-image techniques, AI-based image generation has been applied to various fields.
    However, AI Generated Images (AIGIs) may have some unique distortions compared to natural images, thus many generated images are not qualified for real-world applications. 
    Consequently, it is important and significant to study subjective and objective Image Quality Assessment (IQA) methodologies for AIGIs. 
    In this paper, in order to get a better understanding of the human visual preferences for AIGIs, a large-scale IQA database for AIGC is established, which is named as AIGCIQA2023. We first generate over 2000 images based on 6 state-of-the-art text-to-image generation models using 100 prompts.
    Based on these images, a well-organized subjective experiment is conducted to assess the human visual preferences for each image from three perspectives including \emph{\textbf{quality}}, \emph{\textbf{authenticity}} and \emph{\textbf{correspondence}}. 
    Finally, based on this large-scale database, we conduct a benchmark experiment to evaluate the performance of several state-of-the-art IQA metrics on our constructed database. The AIGCIQA2023 database and benchmark will be released to facilitate future research on \textcolor{magenta!100}{\url{https://github.com/wangjiarui153/AIGCIQA2023}}

\keywords{AI generated content (AIGC) \and text-to-image generation \and image quality assessment \and human visual preference }
\end{abstract}

%
%
\section{Introduction}
Artificial Intelligence Generated Content (AIGC) refers to the content, including texts, images, audios, or videos, \textit{etc.}, that is created or generated with the assistance of AI technology.
Many impressive AIGC models have been developed in recent years, such as ChatGPT 
and DALLE\cite{ramesh2022hierarchical}, which have been utilized in various application scenarios. 
As an important part of AIGC, AI Generated Images (AIGIs) have also gained significant attention in recent years due to advancement in generative models including Generative Adversarial Network (GAN) \cite{goodfellow2020generative}, Variational Autoencoder (VAE) \cite{kingma2013auto}, diffusion models \cite{Rombach2021HighResolutionIS}, \textit{etc.}, and language-image pre-training techniques including CLIP\cite{radford2021learning}, BLIP\cite{li2022blip}, \textit{etc.}

However, the development of AIGI models also raises new problems and challenges. 
One significant challenge is that not all generated images are qualified for real-world applications, which often require to be processed, adjusted, refined or filtered out before being applied to practical scenes.
However, unlike common image content, such as Natural Scene Images (NSIs)\cite{duan2017ivqad,duan2018perceptual}, screen content images\cite{min2017unified,duan2022confusing}, graphic images\cite{min2017unified,duan2022saliency}, \textit{etc.}, which generally encounters some common distortions including  noise, blur, compression, \emph{etc.} \cite{duan2023masked,duan2022develop}, AIGIs may suffer from some unique degradations such as unreal structures, unreasonable combinations, \textit{etc}. Moreover, the generated images may not correspond to the semantics of the text prompts \cite{lee2023aligning,kirstain2023pick,xu2023imagereward}.
Therefore, it is important to study the human visual preferences for AIGIs and design corresponding objective Image Quality Assessment (IQA) metrics for these images.

Many subjective IQA studies have been conducted for human captured or created images, and many objective IQA models have also been developed.
However, these models are designed for assessing low-level distortions, while AIGIs generally contain both low-level artifacts and high-level semantic degradations.
Some quantitative evaluation metrics such as Inception Score (IS)\cite{gulrajani2017improved} and Fréchet Inception Distance (FID)\cite{heusel2017gans} have been proposed to assess the performance of generative models and have been widely used to evaluate the authenticity of the generated images.
However, these methods cannot evaluate the authenticity of a single generated image, and cannot measure the correspondence between the generated images and the text-prompts.
As a new type of image content, previous IQA methods may fail to assess the image quality of AIGIs and cannot align well with human preferences due to the irregular distortions. 

To gain a better understanding of human visual preferences for AIGIs and guide the design process of corresponding objective IQA models, in this paper, we conduct a comprehensive subjective and objective IQA study for AIGIs.
We first establish a large-scale IQA database for AIGIs termed AIGCIQA2023, which contains 2,400 diverse images generated by 6 state-of-the-art AIGI models based on 100 various text prompts. 
Based on these images, a well-organized subjective experiment is conducted to assess the human visual preferences for each individual generated image from three perspectives including 
\textbf{\textit{quality}}, \textbf{\textit{authenticity}}, and \textbf{\textit{correspondence}}. Based on the constructed AIGCIQA2023 database, we evaluate the performance of several state-of-the-art IQA models and establish a new benchmark. Experimental results demonstrate that current IQA methods cannot well align with human visual preferences for AIGIs, and more efforts should be made in this research field in the future. The main contributions of this paper are summarized as follows:

\begin{itemize}

\item We propose to disentangle the human visual experience for AIGIs into three perspectives including \textbf{\textit{quality}}, \textbf{\textit{authenticity}}, and \textbf{\textit{correspondence}}.
\item Based on the above theory, we establish a novel large-scale database, \textit{i.e.,} AIGCIQA2023, to better understand the human visual preferences for AIGIs and guide the design of objective IQA models.
\item We conduct a benchmark experiment to
evaluate the performance of several current state-of-the-art IQA algorithms in measuring the quality, authenticity, and text-image correspondence of AIGIs.

\end{itemize}

The rest of the paper is organized as follows. In Section 2 we introduce the details of our constructed AIGCIQA2023 database, including the generation of AIGIs and the subjective quality assessment methodology and procedures.
In section 3 we present the benchmark experiment for current state-of-the-art IQA algorithms based on the established database. 
Section 4 concludes the whole paper and we discuss possible future research that can be conducted with the database.

\section{Database Construction and Analysis}
In order to get a better understanding of human visual preferences for AI-generated images based on text prompts, we construct a novel IQA database for AIGIs, termed AIGCIQA2023, which is a collection of generated images derived from six state-of-the-art deep generative models based on 100 text prompts, and corresponding subjective quality ratings from three different perspectives.
Then we further analyze the human visual preferences for AIGIs based on the constructed database.

\subsection{AIGI Collection}
\begin{figure}[t]
  \centering
  \includegraphics[width=0.9\textwidth]{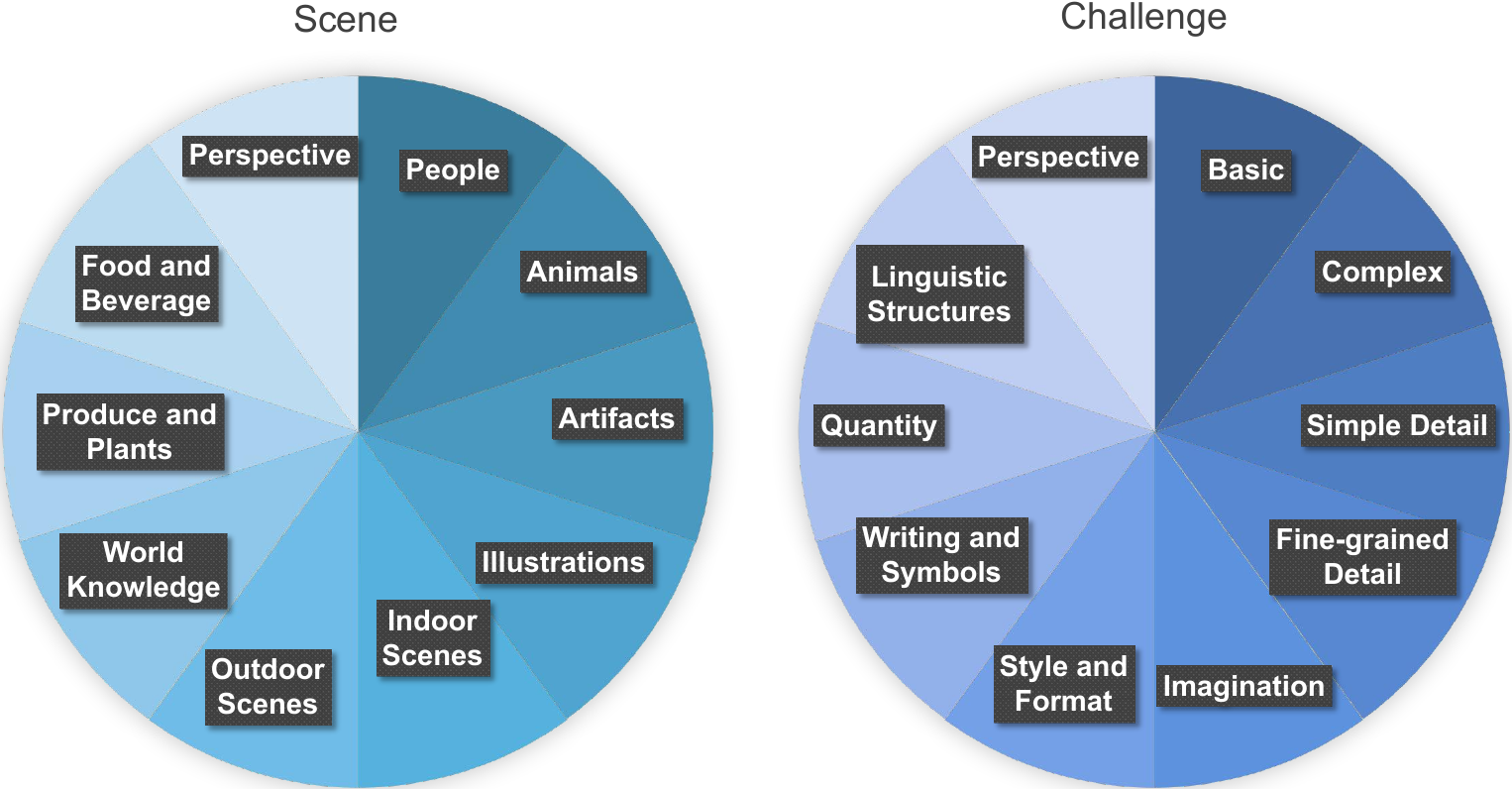}
  \caption{Pie Chart of the ten challenge categories and ten scene categories selected from PartiPrompts\cite{yu2022scaling}. }
\end{figure}
We adopt six latest text-to-image generative models, including Glide\cite{Nichol2021GLIDETP}, Lafite\cite{Zhou_2022_CVPR}, DALLE\cite{ramesh2022hierarchical}, Stable-diffusion\cite{Rombach2021HighResolutionIS}, Unidiffuser\cite{Bao2023OneTF}, Controlnet\cite{Zhang2023AddingCC}, to produce AIGIs by using open source code and default weights.
To ensure content diversity and catch up with the practical application requirements, we collect diverse texts from the PartiPrompts website \cite{yu2022scaling} as the prompts for AIGI generation.
The text prompts can be simple, allowing generative models to produce imaginative results.
They can also be complex, which raises the challenge for generative models.
We select 10 scene categories from the prompt set, and each scene contains 10 challenge categories.
Overall, we collect 100 text prompts (10 scene categories $\times$ 10 challenge categories) from PartiPrompts\cite{yu2022scaling}.
The distribution of the selected scene and challenge categories is displayed in pie chart of Fig.1.
It can be observed that the dataset exhibits a high level of scene diversity, with images generated covering a broad range of challenges.
Then we perform the text-to-image generation based on these models and prompts. Specifically, for each prompt, we generate 4 various images randomly for each generative model. Therefore, the constructed AIGCIQA2023 database totally contains 2400 AIGIs (4 images $\times$ 6 models $\times$ 100 prompts) corresponding to 100 prompts.
\begin{figure}[t]
  \centering
  \includegraphics[width=\textwidth]{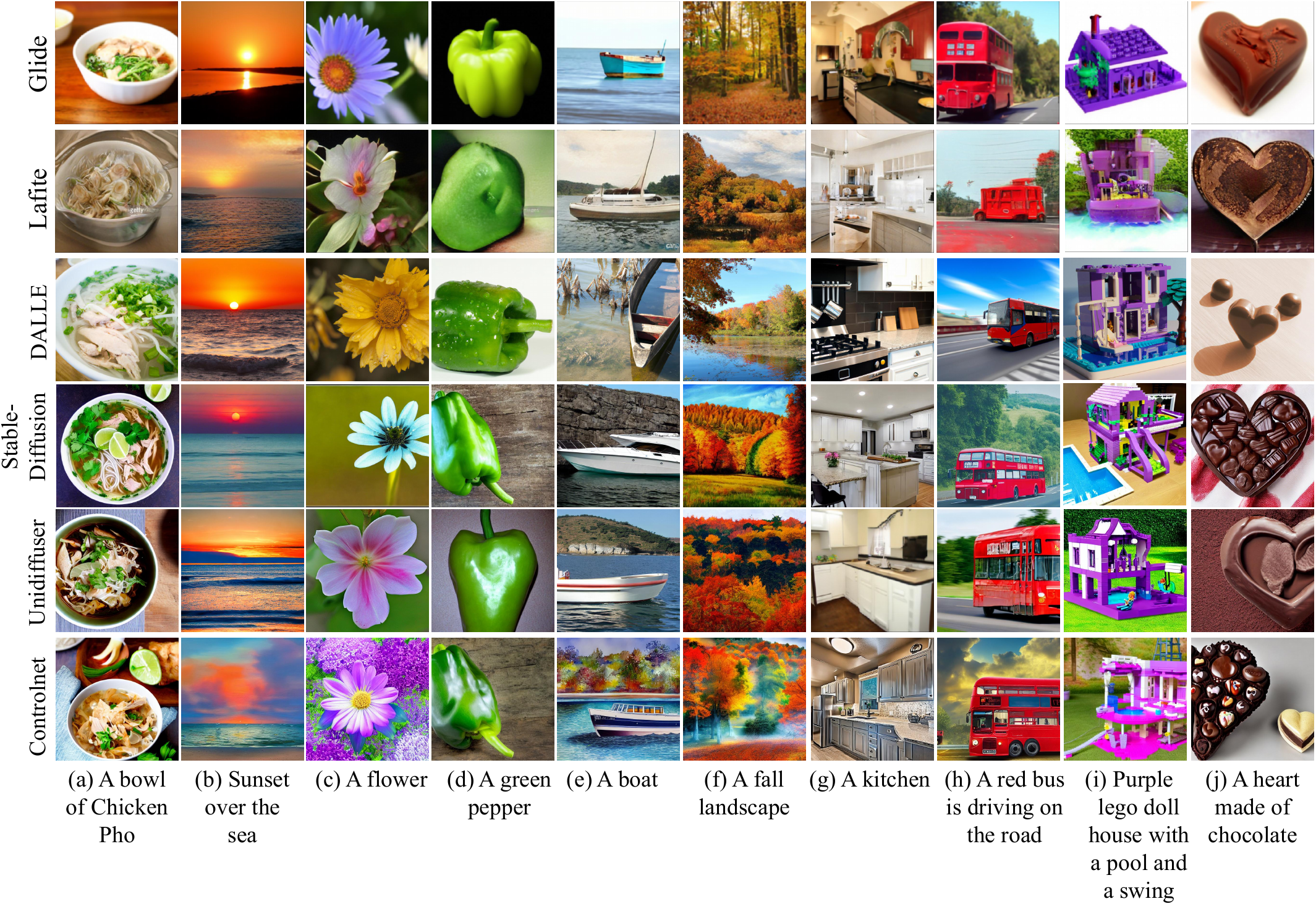}
  \caption{Sample images from the AIGCIQA2023 database generated by six different generative models (Glide\cite{Nichol2021GLIDETP}, Lafite\cite{Zhou_2022_CVPR}, DALLE\cite{ramesh2022hierarchical}, \
  Stable-diffusion\cite{Rombach2021HighResolutionIS}, Unidiffuser\cite{Bao2023OneTF}, Controlnet\cite{Zhang2023AddingCC}.)}
\end{figure}

\subsection{Subjective Experiment Setup}

Subjective IQA is the most reliable way to evaluate the visual quality of digital images perceived by the users. 
It is generally used to construct image quality datasets and served as the ground truth to optimize or evaluate the performance of objective quality assessment metrics. 
Due to the unnatural property of AIGIs and different text prompts having different target image spaces, it is unreasonable to just use one score, \textit{i.e.,} ``quality'' to represent human visual preferences.
In this paper, we propose to measure the human visual preferences of AIGIs from three perspectives including \textbf{\textit{quality}}, \textbf{\textit{authenticity}}, and text-image \textbf{\textit{correspondence}}.
For an image, these three visual perception perspectives are related but different.
\begin{figure}[t]
  \centering
  \includegraphics[width=\textwidth]{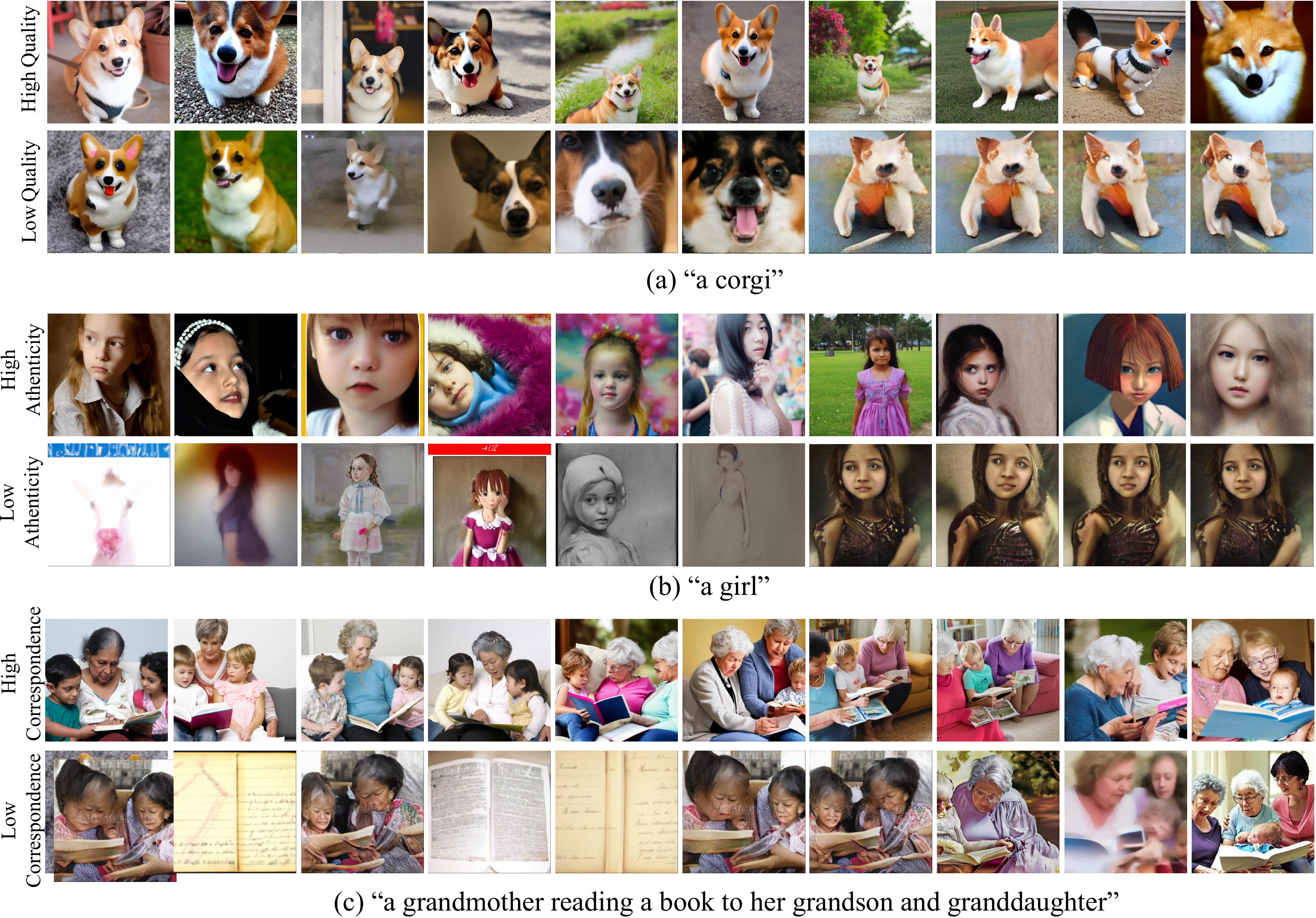}
  \caption{Illustration of the images from the perspectives of quality, authenticity, and text-image correspondence.
  (a)  10 high quality examples
and 10 low quality examples of the images generated by the prompt of “a corgi”.
  (b)  10 high authenticity and
10 low authenticity examples of images generated by the prompt of “a girl”.
  (c)  10 high text-image correspondence and 10 low correspondence
examples of images generated by the prompt of “a grandmother reading a book
to her grandson and granddaughter”.
}

\end{figure}

The first dimension of AIGI evaluation is ``quality'' evaluation, \textit{i.e.,} evaluating an AIGI from its clarity, color, lightness, contrast, \textit{etc.}, which is similar to the assessment of NSIs.
During the experiment procedure, subjects are instructed to evaluate whether the image outline is clear, whether the content can be distinguished, and the richness of details, \textit{etc.} 
Fig.3 (a) shows 10 high quality examples and 10 low quality examples of the images generated by the prompt of “a corgi”.

Considering the generation nature of AIGIs, an important problem of these images is that they may not look real compared to NSIs.
Therefore, we introduce a second dimension of evaluation metrics for the generated images, \textit{i.e.,} ``authenticity'' evaluation.
For this dimension, subjects are instructed to assess the image from the authenticity aspect, \textit{i.e.,} whether it looks real or whether they can distinguish that the image is AI-generated or not. 
Fig.3 (b) shows 10 high authenticity and 10 low authenticity examples of images generated by the prompt of ``a girl''.

Since an AIGI is generated from a text, it is also important to evaluate its correspondence with the original prompt, \textit{i.e.,} the third dimension, text-image ``correspondence''. 
For this purpose, subjects are instructed to consider textual information provided with the image and then give the correspondence score from 0 to 5 to assess the relevance between the generated image and its prompt. 
Fig.3 (c) shows 10 high text-image correspondence and 10 low correspondence examples of images generated by the prompt of “a grandmother reading a book to her grandson and granddaughter''.
\begin{figure}[t]
  \centering
  \includegraphics[width=0.9\textwidth]{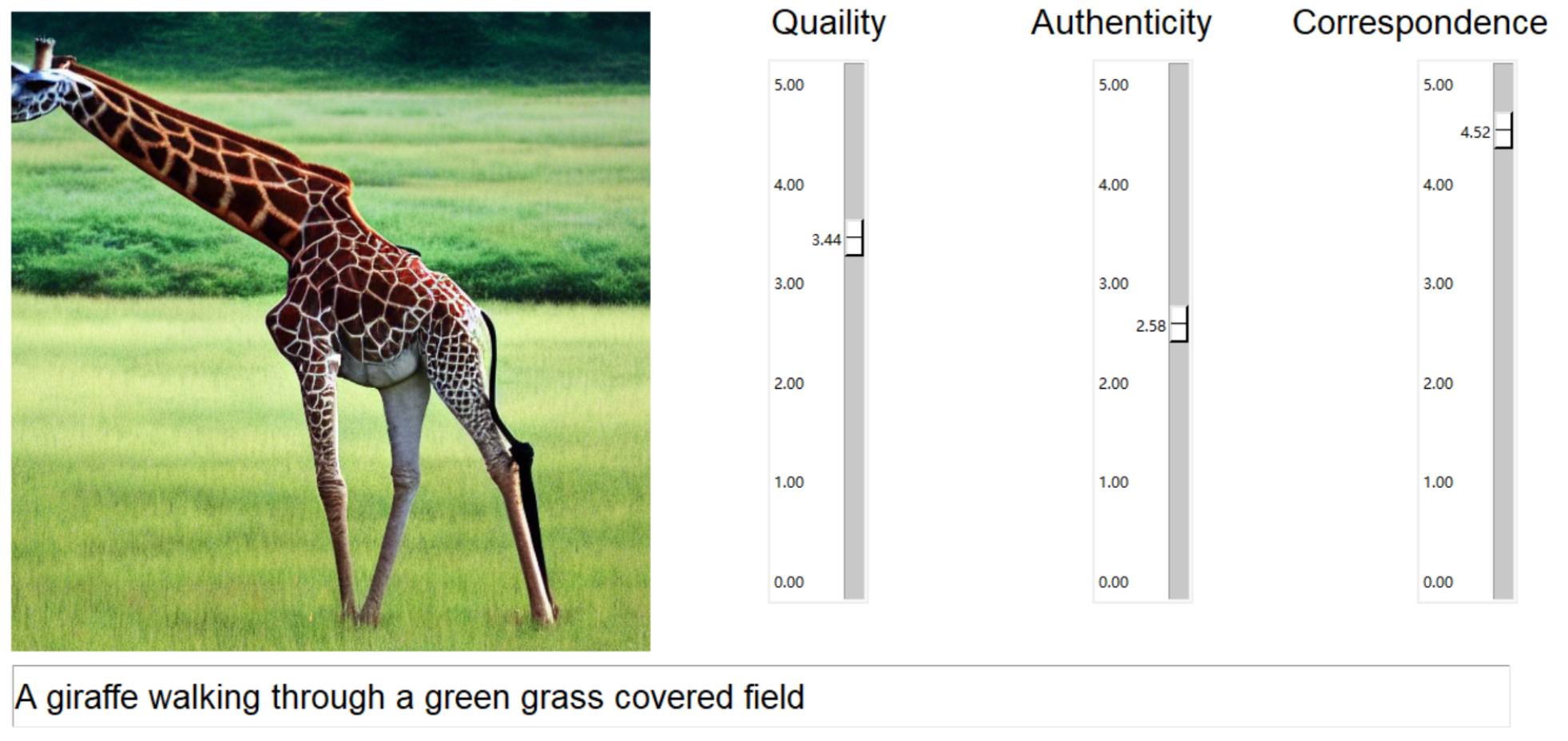}
  \caption{An example of the subjective assessment interface. The subject can  evaluate the quality of AIGIs and record the quality, authenticity, correspondence scores with the scroll bar on the right.}
\end{figure}
\vspace{-2pt}
\subsection{Subjective Experiment Procedure}
To evaluate the quality of the images in the AIGCIQA2023 and obtain Mean Opinion Scores (MOSs), a subjective experiment is conducted following the guidelines of ITU-R BT.500-14 \cite{duan2022confusing}. 
The subjects are asked to rate their visual preference degree of exhibited AIGIs from the quality, authenticity and text-image correspondence.
The AIGIs are presented in a random order on an iMac monitor with a resolution of up to 4096 × 2304, using an interface designed with Python Tkinter, as shown in Fig.4. The interface allows viewers to browse the previous and next AIGIs and rate them using a quality scale that ranges from 0 to 5, with a minimum interval of 0.01. A total of 28 graduate students (14 males and 14 females) participate in the experiment, and they are seated at a distance of around 60 cm in a laboratory  environment with normal indoor lighting.

\subsection{Subjective Data Processing}

We follow the suggestions recommended by ITU to conduct the outlier detection and subject rejection. 
The score rejection rate is 2\%.
In order to obtain the MOS for an AIGI, we first convert the raw ratings into Z-scores, then linearly scale them to the range $[0,100]$ as follows:
$$z_i{}_j=\frac{r_i{}_j-\mu_i{}_j}{\sigma_i},\quad z_{ij}'=\frac{100(z_{ij}+3)}{6},$$
$$\mu_i=\frac{1}{N_i}\sum_{j=1}^{N_i}r_i{}_j, ~~ \sigma_i=\sqrt{\frac{1}{N_i-1}\sum_{j=1}^{N_i}{(r_i{}_j-\mu_i{}_j)^2}}$$ 
where $r_{ij}$ is the raw ratings given by the $i$-th subject to the $j$-th image. $N_i$ is the number of images judged by subject $i$. 

Next, the mean opinion score (MOS) of the image j is computed by averaging the rescaled z-scores as follows:
$$MOS_j=\frac{1}{M}\sum_{i=1}^{M}z_{ij}'$$
where $MOS_j$ indicates the MOS for the $j$-th AIGI, $M$ is the number of valid subjects, and $z'_i{}_j$ are the rescaled z-scores. 
\begin{figure}[t]
    \centering
    \includegraphics[width=\textwidth]{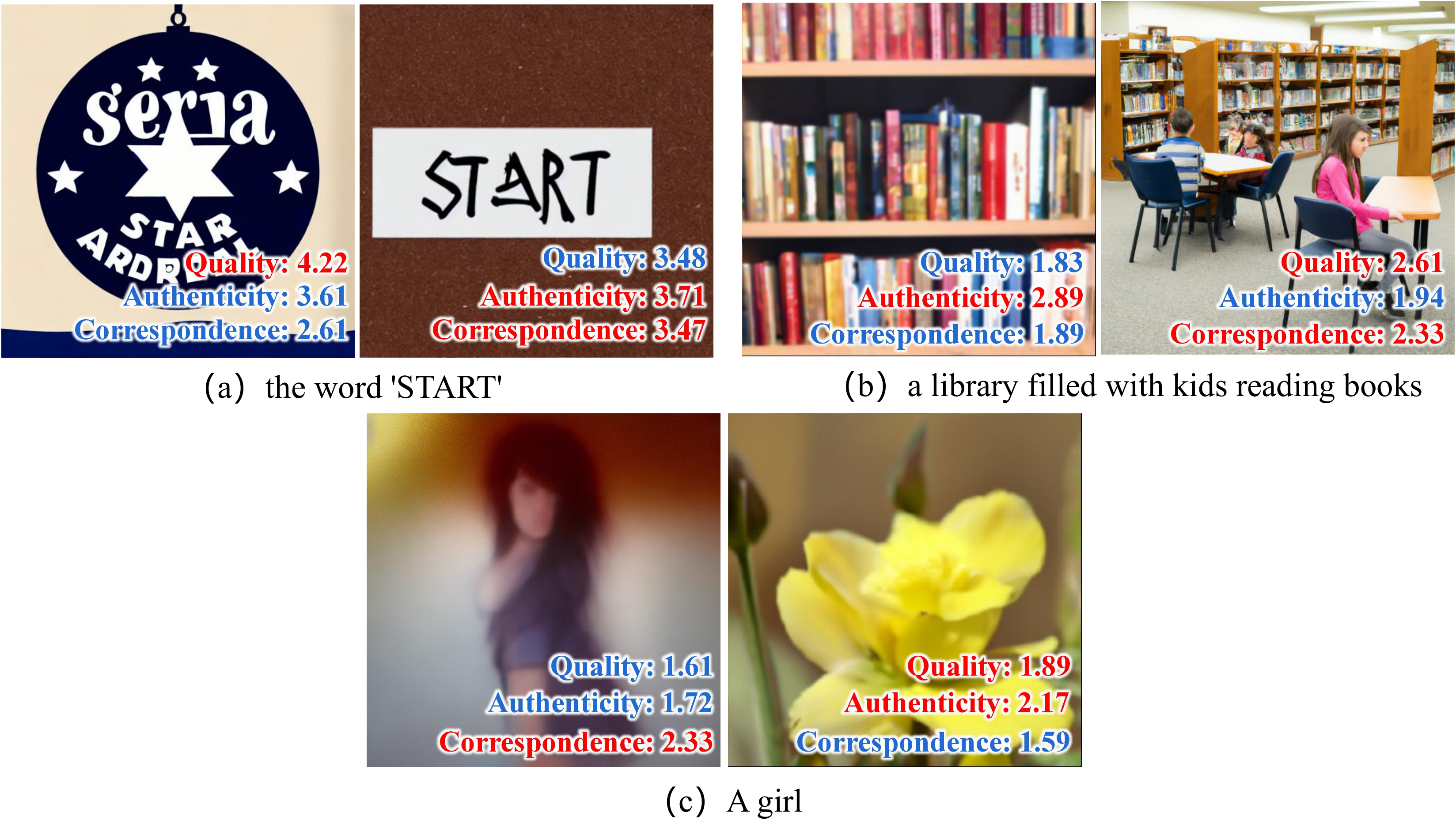}
    \caption{Comparison of the differences between three evaluation perspectives. (a) Left image has better quality, but worse authenticity and correspondence. (b) Left image has better authenticity, but worse quality and correspondence. (c) Left image has better correspondence, but worse quality and authenticity.}
    \label{fig:my_label}
\end{figure}

\subsection{AIGI Analysis from Three Perspectives}
\vspace{-2pt}

To further illustrate the differences of the three perspectives, we demonstrate several example images and their corresponding subjective ratings from three aspects in Fig.5.
For each subfigure, it can be noticed that the right AIGI outperforms the left AIGI on two evaluation dimensions but is much worse than the left AIGI on another dimension, which demonstrates that each evaluation perspective (quality, authenticity, or text-image correspondence) has its own unique perspective and value.

\begin{figure}[t]
  \centering
  \includegraphics[width=\textwidth]{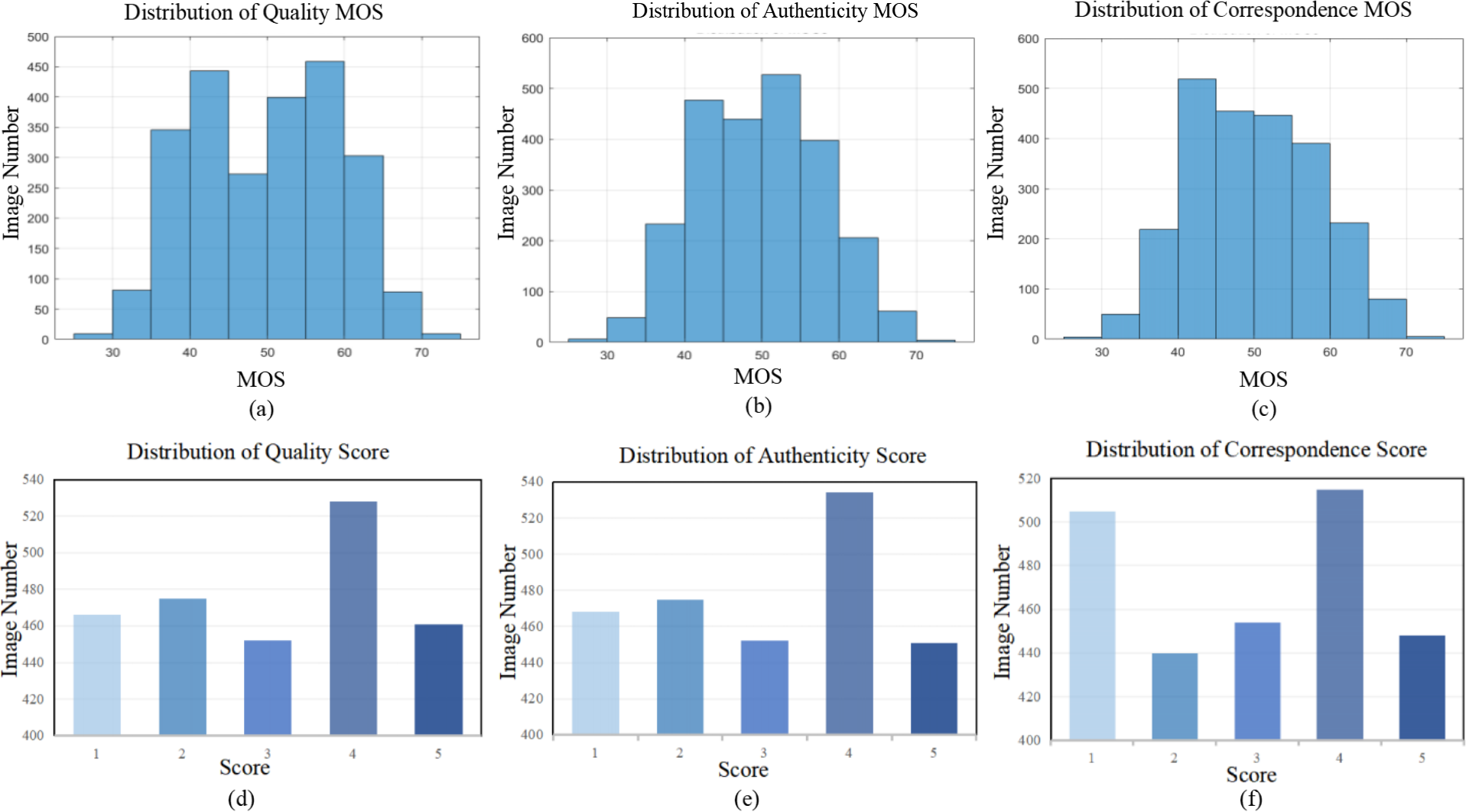}
  \caption{ (a) MOSs distribution of quality score.
  (b) MOSs distribution of authenticity score.
  (c) MOSs distribution of correspondence score.
  (d) Distribution of the quality score.
  (e) Distribution of the authenticity score.
  (f) Distribution of the correspondence score.}
\end{figure}
Fig.6 demonstrates the MOS and score distribution for quality evaluation, authenticity evaluation, and text-image correspondence evaluation, respectively, which demonstrate the images in AIGCIQA 2023 cover a wide range of perceptual quality. 

\section{EXPERIMENT}
\subsection{Benchmark Models}
Since the AIGIs in the proposed AIGCIQA2023 database are generated based on text prompts and have no pristine reference images, they can only be evaluated by no-reference (NR) IQA metrics.
In this paper, we select fifteen state-of-the-art IQA models for comparison. The selected models can be classified into two groups:

\begin{itemize}
\item \textbf{Handcrafted-based} models, including: NIQE\cite{mittal2012making}, BMPRI\cite{min2018blind}, BPRI\cite{min2017blind}, BRISQUE\cite{mittal2012no}, HOSA\cite{xu2016blind}, BPRI-LSSn\cite{min2017blind}, BPRI-LSSs\cite{min2017blind}, BPRI-PSS\cite{min2017blind}, QAC\cite{xue2013learning}, HIGRADE-1 and HIGRADE-2\cite{kundu2017large}. 

These models extract handcrafted features based on prior knowledge about image quality. 

\item \textbf{Deep learning-based} models, including: CNNIQA\cite{kang2014convolutional}, WaDIQaM-NR\cite{bosse2017deep}, VGG (VGG-16 and VGG-19)\cite{simonyan2014very} and ResNet (ResNet-18 and ResNet-34)\cite{he2016deep}.  

These models characterize quality-aware information by training deep neural networks from labeled data.
\end {itemize}

\subsection{Evaluation Criteria}

In this study, we utilize the following four performance evaluation criteria to evaluate
the consistency between the predicted scores and the corresponding ground-truth MOSs, including Spearman Rank Correlation Coefficient (SRCC), Pearson Linear Correlation Coefficient (PLCC), Kendall’s Rank Correlation Coefficient (KRCC), and Root Mean Squared Error (RMSE).


\subsection{ Experimental Setup}
All the benchmark models are validated on the proposed AIGCIQA2023 database. 
For traditional handcrafted-based models, they are directly evaluated based on the database.
For deep trainable models, we first randomly split the database into an 4:1 ratio for training/testing while ensuring the image with the same prompt label falls into the same set. 
The partitioning and evaluation process is repeated several times for a fair comparison while considering the computational complexity, and the average result is reported as the final performance. 
For deep learning-based models, we applied CNNIQA\cite{kang2014convolutional}, WaDIQaM-NR\cite{bosse2017deep}, VGG (VGG-16 and VGG-19)\cite{simonyan2014very} and ResNet (ResNet-18 and ResNet-34)\cite{he2016deep} to predict the MOS of image quality. 
The repeating time is 10, the training epochs are 50 with an initial learning rate of 0.0001 and batch size of 4.

\subsection{Performance Discussion}
\begin{table}

\caption{Performance comparisons of the state-of-the-art IQA methods on the AIGCIQA2023 database. The best performance results are marked in \textcolor{red}{RED} and the second-best performance results are marked in \textcolor{blue}{BLUE} .}
  \label{t2}
  \centering
  \resizebox{\textwidth}{!}{
  \begin{tabular}{l||c|c|c|c|c|c|c|c|c|c|c|c}
    \toprule
    \multicolumn{1}{l||}{} &
    \multicolumn{4}{|c|}{\textbf{Quality}} &
    \multicolumn{4}{|c|}{\textbf{Authenticity}} &
    \multicolumn{4}{|c}{\textbf{Correspondence}} 
    \\
    \cmidrule(r){1-13}
    
    \textbf{Method}  & \textbf{SRCC}  & \textbf{KRCC}  & \textbf{PLCC}   & \textbf{RMSE}  & \textbf{SRCC}  & \textbf{KRCC}  & \textbf{PLCC}   & \textbf{RMSE} & \textbf{SRCC}  & \textbf{KRCC}  & \textbf{PLCC}   & \textbf{RMSE} \\
    \midrule

\textbf{NIQE}\cite{mittal2012making} & 0.5060 & 0.3420 & 0.5218 & 7.9461 &0.3715 &0.2453 & 0.3954 & 7.3999 & 0.3659 &0.2460 &0.3485 
&7.7721    \\
  \textbf{QAC}\cite{xue2013learning} &0.5328 &0.3644 &0.5991 &6.3062 
&0.4009 &0.2673 &0.4428 &7.2236
&0.3526 &0.2414 &0.4062 &7.5768   \\
    \textbf{BRISQUE}\cite{mittal2012no} &0.6239 &0.4291 &0.6389 &7.1655 
&0.4705 &	0.3142 &0.4796 &	7.0695  
&0.4219 &	0.2865 &0.4280 & 7.4941    \\
 \textbf{PRI-PSS}\cite{min2017blind} &0.3556 &0.2373 &0.4183 &8.4605 
&0.2409 &0.1583 &0.2625 &7.7739
&0.2670 &0.1794 &0.2960 &7.9203  \\
    \textbf{PRI-LSSs}\cite{min2017blind} &0.5141 &0.3512 &0.5618 &7.7054 
&0.3721 &0.2460 &0.3998 &7.3845  
&0.3230 &0.2160 &	0.3473 &7.7756   \\
    \textbf{PRI-LSSn}\cite{min2017blind} &0.5245 &0.3523 &0.5935 &7.4964 
&0.3838 &0.2528 &0.5465 &6.7467  
&0.3655 &0.2474 &0.4594 &7.3653   \\
   \textbf{BPRI} \cite{min2017blind} &0.6301 &	0.4307 &0.6889 &6.7517 
&0.4740 &	0.3144 &0.5207& 6.8783  
&0.3946 &	0.2657 &0.4346&7.4680   \\
 
    \textbf{HOSA}\cite{xu2016blind} &0.6317 &0.4311 &	0.6561&	7.0297 
&0.4716 &0.3101 &0.4985 &6.9841 
&0.4101 &0.2765 &0.4252 &7.5051   \\
\textbf{BMPRI}\cite{min2018blind} &0.6732 &0.4661 &0.7492 &6.1693 
&0.5273 &	0.3554 &0.5756&6.5878 
&0.4419 &	0.3014 &0.4827 &7.2619  \\

\textbf{Higrade-1}\cite{kundu2017large} &0.4849 &0.3220&	0.4966&	8.0847 
&0.4175& 	0.2791& 	0.4181& 	7.3183 
&0.3319& 	0.2207& 	0.3379& 	7.8041  \\
\textbf{Higrade-2} \cite{kundu2017large}&0.2344 &	0.1568& 0.3189& 	8.8282 
&0.2654 &	0.1742 &	0.3106 &	7.6579 
&0.1756&0.1170 &	0.2144&8.0990  \\
\cmidrule(r){1-13}

\textbf{WaDIQaM-NR}\cite{bosse2017deep}& 0.4447 &	0.3036 &	0.4996 &	8.7400 
&0.3936 &	0.2715 &	0.3906 &	7.4627 
&0.3027 &0.2057 &	0.2810 &	\textcolor{red}{6.0477}
 \\
 \textbf{CNNIQA}\cite{kang2014convolutional} &0.7160 &0.4955 &0.7937 &\textcolor{blue}{5.8816}
&0.5958 &0.4085&0.5734 &6.7231  
&0.4758 &0.3313&	0.4937 &	7.3839   \\
 \textbf{VGG16}\cite{simonyan2014very} &\textcolor{red}{0.7961}  &\textcolor{red}{0.5843}  &	\textcolor{blue}{0.7973} &	6.2143  
&0.6660  &\textcolor{blue}{0.4813}  &\textcolor{red}{0.6807}  &\textcolor{red}{6.0273} 
&\textcolor{blue}{0.6580}  &\textcolor{blue}{0.4548}  &\textcolor{blue}{0.6417}  &6.9292    \\
 \textbf{VGG19}\cite{simonyan2014very} &\textcolor{blue}{0.7733}   &\textcolor{blue}{0.5376}   &	\textcolor{red}{0.8402}  &	\textcolor{red}{5.0860}   
&\textcolor{blue}{0.6674}  &\textcolor{red}{0.4843}   &\textcolor{blue}{0.6565}   &\textcolor{blue}{6.1705} 
&0.5799  &0.4090   &0.5670   &6.9851     \\
 \textbf{Resnet18}\cite{he2016deep} &0.7583    &0.5360    &	0.7763   &	6.9897    
&\textcolor{red}{0.6701}   &0.4740    &0.6528    &6.4597   
&0.5979  &0.4165  &0.5564   &7.0957     \\
 \textbf{Resnet34}\cite{he2016deep} &0.7229     &0.4835     &	0.7578    &	6.4806     
&0.5998    &0.4325     &0.6285     &6.5344    
&\textcolor{red}{0.7058}   &\textcolor{red}{0.5111}   &\textcolor{red}{0.7153}   &\textcolor{blue}{6.7605}      \\
    \bottomrule
  \end{tabular}
  }
\end{table}
The performance results of the state-of-the-art IQA models mentioned above on the proposed AIGCIQA2023 database are exhibited in Table 1, from which we can make several conclusions:
\begin{itemize}
\item The handcrafted-based methods achieve poor performance on the whole database, which indicates the extracted handcrafted features are not effective for modeling the quality representation of AIGIs. This is because most employed handcrafted features of
these methods are based on the prior knowledge learned from NSIs, which are not effective for evaluating AIGIs.

\item The deep learning-based methods achieve relatively more competitive performance results on three evaluation perspectives. However, they are still far away from satisfactory.

\item Most of the IQA models achieve better performance on quality evaluation and worse on text-image correspondence score assessment.
The reason is that the text prompts for image generation are not utilized for the IQA model training. 
This makes it more challenging for the IQA models to extract relation features from AIGIs, which inevitably leads to performance drops.
\end {itemize}

\section{Conclusion and Future Work}
In this paper, we study the human visual preference problem for AIGIs. 
We first construct a new IQA database for AIGIs, termed AIGCIQA2023, which includes 2400 AIGIs generated based on 100 various text-prompts, and corresponding subjective MOSs evaluated from three perspectives (\textit{i.e., quality, authenticity, and text-image correspondence}).
Experimental analysis demonstrates that these three dimensions can reflect different aspects of human visual preferences on AIGIs, which further manifests that the evaluation of Quality of Experience (QoE) for AIGIs should be considered from multiple dimensions.
Based on the constructed database, we evaluate the performance of several state-of-the-art IQA models and establish a new benchmark to facilitate future research.

In future work, we will further explore the human visual perception for AIGIs and develop corresponding objective evaluation models for better assessing the quality of AIGIs from the three perspectives proposed in this paper.

\newpage

\bibliographystyle{splncs04}
\bibliography{ref.bib}
\end{document}